\documentclass{article}
\usepackage[margin=1in]{geometry}
\usepackage[utf8]{inputenc}
\usepackage{float,graphicx}
\newcommand{\calS}{\mathcal{S}}
\usepackage{url,hyperref,subcaption}
\usepackage{lipsum}
\usepackage{authblk}
\usepackage{algorithm}
\usepackage{algorithmic}

\usepackage{amssymb, amsmath, mathtools}
\usepackage[flushleft]{threeparttable}
\usepackage{booktabs,multirow,makecell}

\usepackage{natbib}
 \bibpunct[, ]{(}{)}{,}{a}{}{,}%

\usepackage{algorithm}
\usepackage{algorithmic}

\usepackage{amssymb, amsmath, amsthm, mathtools}

\begin{document}

\title{Multi-Agent Reinforcement Learning for \\Joint Police Patrol and Dispatch}
\author{Matthew Repasky\thanks{Email: mwrepasky@gatech.edu}}
\author{He Wang}
\author{Yao Xie}
\affil{
\small
H. Milton Stewart School of Industrial and Systems Engineering, Georgia Institute of Technology}
\date{\vspace{-20pt}}

\maketitle

\begin{abstract}
Police patrol units need to split their time between performing preventive patrol and being dispatched to serve emergency incidents. In the existing literature, patrol and dispatch decisions are often studied separately. We consider joint optimization of these two decisions to improve police operations efficiency and reduce response time to emergency calls.
\textit{Methodology/results:} We propose a novel method for jointly optimizing multi-agent patrol and dispatch to learn policies yielding rapid response times. Our method treats each patroller as an independent \texorpdfstring{$Q$}{Q}-learner (agent) with a shared deep \texorpdfstring{$Q$}{Q}-network that represents the state-action values. The dispatching decisions are chosen using mixed-integer programming and value function approximation from combinatorial action spaces. We demonstrate that this heterogeneous multi-agent reinforcement learning approach is capable of learning joint policies that outperform those optimized for patrol or dispatch alone. 
\textit{Managerial Implications:} Policies jointly optimized for patrol and dispatch can lead to more effective service while targeting demonstrably flexible objectives, such as those encouraging efficiency and equity in response.
\end{abstract}

\maketitle

\section{Introduction}\label{sec:intro}
The timeliness of police response to emergency calls plays a critical role in maintaining public safety and preventing crimes. 
However, recent data showed that police response times in major U.S. cities are getting longer, partly due to staffing shortages \citep{Kaste2023}. In this paper, we study how to improve police patrol operations and reduce emergency call response times by multi-agent reinforcement learning, where each patrol unit (e.g., a police patrol car) is treated as an agent.

Coordinating the patrol of multiple agents in a shared environment has been a task of interest in operations research and machine learning for many years. Approaches ranging from extensions of the traveling salesman problem \citep{chevaleyre2004theoretical} to deep reinforcement learning (RL) for multi-agent robot patrol \citep{jana2022deep} have been applied. Many challenges are associated with multi-agent patrol, including large action spaces that grow exponentially with the number of patrollers, coordination/cooperation of agents, and state representation.
Some prior works evaluate patrol policies using criteria that capture ``coverage,'' such that all locations are visited frequently \citep{almeida2004recent}. Instead, in this work, we evaluate patrol policies with respect to a separate but related task: multi-agent dispatch.

\begin{figure}[b!]
    \centering
    \includegraphics[width=\textwidth]{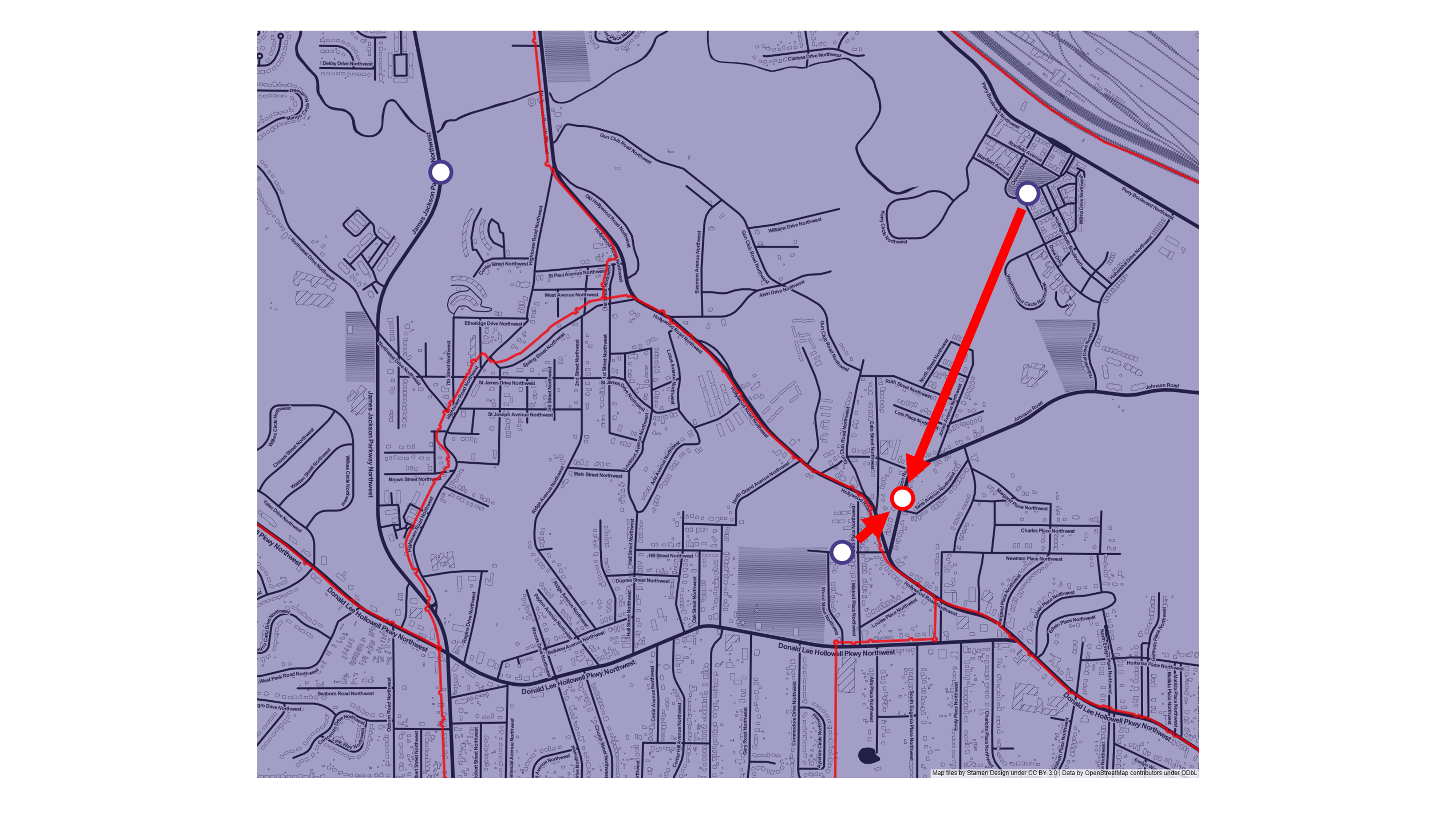}
    \caption{The joint task of patrol and dispatch involves decisions such as that demonstrated above. It is unclear whether it is best to move one patroller across beat boundaries to manage an incident, leaving its beat un-managed, or to require the patroller responsible for that beat to travel a long distance to the scene.}
    \label{fig:intro}
\end{figure}

The police dispatching problem has traditionally been considered using multi-server priority queues \citep{green1984multiple}. The task of a dispatcher is to select a patroller to dispatch to the scene of an incident. Recent efforts have considered the effects of dispatch policies given positioning and/or routing of patrollers in some space (e.g., a grid or a graph) \citep{guedes2015multi,dunnett2019optimising}. Oftentimes, the response times to incidents must meet some key performance indicators \citep{stanford2014waiting}, which has also been considered in the context of dispatch \citep{zhang2016simulation}.

The positioning of patrollers is key to effective dispatch, yet most dispatch works consider heuristic patrol policies. Early works considering multi-agent patrol (without dispatch) coordinated individual agent patrol decisions using heuristics such as neighboring node idleness \citep{machado2003multi}. Others considered extensions of the single-patroller traveling salesman problem to multiple agents \citep{chevaleyre2004theoretical}. Multi-agent reinforcement learning (MARL) for multi-agent patrolling has also been considered for many years \citep{santana2004multi}; e.g., node idleness can be minimized by independent learners with cooperative rewards. Early RL approaches considered individual patrollers who only had information about their local environment but could communicate with others about intended actions. More recently, multi-agent patrol policies have been learned using Bayesian learning \citep{portugal2013applying} and deep RL \citep{jana2022deep}. A common theme is distributed policy optimization with varying degrees of communication between agents, where the goal is to optimize with respect to idleness or frequency of visits to each location.

{In addition to a growing need for efficient police operations in light of staffing shortages and limited resources in recent years~\citep{apnews2023hiring,charalambous2023vicious,hermann2023DC}, fairness and equity in operations such as patrol and dispatch is an area of increased interest, and the benefits of such considerations are plentiful. For instance, \cite{dickson2022identifying} shows that police officers who are perceived to be fair are more likely to observe respect and legitimacy with their community. Moreover, opinions within the US justice system echo the sentiment that fairness in law enforcement is crucial for public cooperation and support~\citep{waldschmidt2018perspective}. The importance of this issue is highlighted by many police departments across the US, often posting statements and procedures regarding fairness in their operations~\citep{parkerpolice,notredamepolice,walker2023importance}. Algorithmic considerations of fairness in police operations typically focus on providing equity in service to multiple groups~\citep{wheeler2020allocating,liberatore2023towards}.}

To our knowledge, ours is the first work to optimize policies for patrol and dispatch jointly. Our contributions are as follows.
\begin{itemize}
    
    \item\textbf{Jointly-optimized policies for patrol and dispatch}: Prior dispatch optimization strategies typically assume heuristic patrol policies. Similarly, prior multi-agent patrolling works do not jointly optimize separate decision tasks such as dispatch. We outline a procedure to optimize policies for patrol and dispatch jointly, outperforming benchmark policies optimized for patrol or dispatch alone in addition to heuristic policies such as priority queue dispatch. We demonstrate that the superior performance of our method is robust to a range of simulated environments of varying sizes and dynamics and with different objectives.

    \item\textbf{Novel optimization approach}: We use distributed MARL with parameter sharing to exploit similarities between patrollers while jointly optimizing a dispatcher agent whose action space has combinatorial cardinality. We extend recent works in combinatorial RL to systems with stochastic transitions. We incorporate them into our MARL framework via a coordinate-descent-style alternating optimization of patrol and dispatch policies.

    \item\textbf{Fairness and real service systems}: We demonstrate that our approach is flexible to various reward definitions, such as those that encourage equitable policies. Policies that explicitly encode equity in the reward definition yield more balanced responses to groups of different sizes. We also apply our approach to scenarios based on real calls-for-service data.

\end{itemize}

\subsection{Related works}\label{sec:related}
RL is a machine learning approach for training an agent to make sequential decisions in a stochastic environment \citep{sutton2018reinforcement,bertsekas2019reinforcement}. Deep RL uses neural networks, such as deep $Q$ networks (DQNs), to learn policies \citep{mnih2015human}. Making decisions for multiple agents is a central issue in MARL, as the number of possible actions is exponential in the number of agents. Independent $Q$-learning (IQL) \citep{tan1993multi} addresses this by decomposing the problem into multiple single-agent RL problems. Each agent operates in the same environment, but there are no guarantees regarding cooperation and interference; nonetheless, the naive extension of DQNs to the IQL framework has been successful in some cases \citep{gupta2017cooperative,tampuu2017multiagent}. To address cooperation and efficiency in distributed MARL, parameter sharing in deep RL has been employed \citep{gupta2017cooperative,chu2017parameter,terry2020parameter,christianos2021scaling,chen2021deep,yu2022dinno}. Greater degrees of parameter sharing are often most effective between agents with similar value spaces and reward functions.

Multi-agent patrol has been investigated from the perspective of operations research \citep{chevaleyre2004theoretical}, Bayesian learning \citep{portugal2013applying}, and (deep) MARL \citep{santana2004multi,jana2022deep}. Many of these works pose the problem as a discrete-time, discrete-location graph traversal \citep{almeida2004recent,portugal2010msp,portugal2013applying}. Past works recognize the inherent complexity of making a centralized decision to patrol many agents and decompose the problem via distributed optimization \citep{santana2004multi, farinelli2017distributed}. Similarly, our approach to learning multi-agent patrol policies is distributed while addressing issues arising from the decentralized decision-making for multiple cooperative agents. Additionally, we consider this problem in conjunction with a separate but connected dispatch task, which is not thoroughly addressed in prior work.

Dispatch of multiple patrollers and resources is typically considered from the queueing or ranking systems perspective. \cite{zhang2016simulation} investigate such a setting using static and dynamic priority queues, connecting to the queuing literature in developing dispatch policies and indicating the static queues yield better performance. For instance, \cite{guedes2015multi} construct a ranking system based on quantitative incident characteristics, such as arrival time, priority, and distance to the nearest patroller. Similarly, \cite{dunnett2019optimising} formulates the dispatch decision process by constructing multiple queues, considering incident categories and severity, amongst other factors such as routing of units. Crucially, prior works regarding dispatch in service systems neglect to optimize multi-agent patrol policies, typically assuming basic heuristics.

A typical notion of fairness in machine learning applications suggests that the outcome of an algorithm should be independent of the group to which it is applied~\citep{barocas2017fairness}. In the context of service operations such as police patrol and dispatch, such groups can be defined by socioeconomic or racial status. For example, fairness has been studied in the police districting problem to improve efficiency while mitigating racial disparity in defining patrol districts~\citep{liberatore2022equity,liberatore2023towards}. Parallel works have worked to alleviate disproportionate minority contact that can arise due to hot spot policing~\citep{wheeler2020allocating}. Such works find that simply neglecting to take sensitive features (e.g., race or location) into account is ineffective at achieving fairnesss~\citep{eckhouse2019layers}. Therefore, these attributes are taken into account in decision-making problems to encode fairness explicitly, often leading to a minimal tradeoff in efficiency.

\section{Background}\label{sec:background}

This section provides the necessary background information regarding reinforcement learning (RL) and multi-agent reinforcement learning (MARL) as applied to service systems such as police patrol and dispatch.

\subsection{Single-Agent Reinforcement Learning}\label{sec:single_agent_rl}

An infinite-horizon discounted Markov Decision Process (MDP) can be represented by a tuple $\langle \mathcal{S}, \mathcal{A}, \mathbb{P}, R, \gamma \rangle$ with state space $\mathcal{S}$, action space $\mathcal{A}$, transition dynamics $\mathbb{P}:\mathcal{S}\times\mathcal{A}\rightarrow \Delta(\mathcal{S})$, reward function $R:\mathcal{S}\times\mathcal{A}\times\mathcal{S}\rightarrow\mathbb{R}$, and discount $\gamma\in[0,1)$. The goal of RL \citep{bertsekas2019reinforcement} is to learn a \textit{policy} $\pi:\mathcal{S}\rightarrow\mathcal{A}$ maximizing cumulative discounted reward, captured by the \textit{value function} $V^{\pi}:\mathcal{S}\rightarrow\mathbb{R}$:
\begin{equation*}
    V^{\pi}(s_0) = \mathbb{E} \left[\sum_{t\geq0} \gamma^t R\left(s_t, \pi(s_t), s_{t+1} \right) \right],
\end{equation*}
beginning in state $s_0$ and following $a_t \sim \pi(s_t)$, where $s_t$ is the state in period $t$. A policy that maximizes the value function $V^{\pi}(s_0)$ above is called an optimal policy and is denoted by $\pi^{\star}$. Let $V^{\star}$ be the optimal value function.

It is often more convenient to consider the \textit{state-action value} ($Q$-value) in RL:
\begin{equation*}
    Q^{\pi}(s_0, a_0) = \mathbb{E} \left[ R(s_0, a_0, s_1) + \sum_{t\geq1} \gamma^t R\left(s_t, \pi(s_t), s_{t+1} \right) \right].
\end{equation*}
The state value $V^{\pi}(s)$ and the state-action value $Q^{\pi}(s, a)$ are connected by $V^{\pi}(s) = Q^{\pi}(s,\pi(s))$. Let $Q^{\star}$ be the optimal Q-value. 
The optimal policy $\pi^{\star}$ takes greedy actions with respect to $Q^{\star}$: $\pi^{\star}(s) = \arg\max_{a \in \mathcal{A}}Q^{\star}(s, a)$.
Since $V^{\star}=\max_a Q^{\star}(s,a)$, the Bellman optimality equation reveals:
\begin{equation*}
    Q^{\star}(s, a) = \mathbb{E} \left[ R(s, a, s') +  \gamma V^{\star}(s') \right].
\end{equation*}

\subsubsection{Deep \texorpdfstring{$Q$}{Q}-Learning.}\label{sec:qlearning}
$Q$-learning is a model-free RL technique to learn the optimal state-action value function by exploiting the Bellman optimality condition \citep{watkins1989learning}. $Q^\pi$, approximated by a function parameterized by $\theta$, can be learned given transitions $(s_t, a_t, s_{t+1}, r_t)\in\mathcal{D}$, where $r_t := R(s_t, a_t, s_{t+1})$ and $\mathcal{D}$ sampled from the MDP. Parameter $\theta$ can then be learned by minimizing the least-squares value iteration (LSVI) objective \citep{bradtke1996linear}:
\begin{equation*}
    \hat{\theta} = \underset{\theta}{\rm argmin} \left( \sum_{(s_t, a_t, s_{t+1}, r_t)\in\mathcal{D}_{\ell}\subset\mathcal{D}} \left( r_t + \gamma\ \underset{a}{\rm max} Q^{\tilde{\theta}}(s_{t+1}, a) - Q^{\theta}(s_t, a_t) \right)^2 \right),
\end{equation*}
A deep Q-network (DQN) \citep{mnih2015human} $Q^\theta: \mathbb{R}^d \rightarrow \mathbb{R}^{|\mathcal{A}|}$ is often used as the approximator, taking $s\in\mathcal{S}\subset\mathbb{R}^d$ and outputting $Q^{\theta}(s,a)$ values for each action. \textit{Target networks} parameterized by $\tilde{\theta}$, a copy of $\theta$ cloned every $L$ updates, are often used to stabilize optimization.

\subsubsection{Value Function Approximation.}\label{sec:value_learning}
One might instead want to approximate $V^\pi(s)$ given policy $\pi$. Assume tuples $(s_t,\tilde{r}_t)\in\mathcal{E}$ are observed, where $\tilde{r}_t := \sum_{t'\geq t} \gamma^{t'-t} R(s_{t'},\pi(s_{t'}), s_{t'+1})$ is the discounted sum of reward following state $s_t$ under policy $\pi$. Value $V^\pi(s)$ can be approximated by a function $V^\phi$ parameterized by $\phi$ to minimize the following objective:
\begin{equation*}
    \hat{\phi} = \underset{\phi}{\rm argmin} \left( \sum_{(s_t, \tilde{r}_t)\in\mathcal{E}} \left(  V^{\phi}(s_t) - \tilde{r}_t \right)^2 \right).
\end{equation*}

\subsubsection{Reinforcement Learning with Combinatorial Action Space}\label{sec:combinatorial_rl}
If $|\mathcal{A}|$ is too large to enumerate (e.g., selection amongst combinatorially-many sets of pairings), $Q$-learning is not applicable since it must enumerate all action values. Let $x_{ij}=1$ if $i$ is paired to $j$, with $x_{ij}=0$ otherwise. Define $d_{ij}$ the reward for this pairing. Since the optimal action maximizes the current reward plus the next state value \citep{delarue2020reinforcement}:
\begin{equation*}
    \pi(s) = \underset{\{x_{ij}\}}{\rm argmax} \ \sum_i
    \sum_j d_{ij}x_{ij} + V^\pi(s'),
\end{equation*}
subject to problem-specific constraints. If $s'$ can be computed given the action and $V^\pi$ is approximated as in Section~\ref{sec:value_learning}, actions are selected by solving a mixed-integer program (MIP).

\subsection{Multi-Agent Reinforcement Learning}\label{sec:marl}

Given agents $i=1,\dots,N$, centralized MARL formulates decision-making as an MDP \citep{gupta2017cooperative}; one agent learns a policy $\pi$ mapping states to a joint action in $\mathcal{A}_1\times\cdots\times\mathcal{A}_N$. IQL addresses the exponential growth of this joint action space by learning $N$ functions $Q^{\theta_i}$ \citep{tan1993multi}. Agent $i$ operates within the single-agent MDP represented by the tuple $\langle \mathcal{S}, \mathcal{A}_i, \mathbb{P}_i, R_i, \gamma_i \rangle$. When the agents are similar, parameter sharing can accelerate learning, acting as an intermediate between centralized MARL and IQL \citep{gupta2017cooperative}.

MARL can also be described using a Markov game (MG) \citep{littman1994markov} formulation, defined by a tuple $\langle N, \mathcal{S}, \{\mathcal{S}_i\}_{i=1}^{N}, \{\mathcal{A}_i\}_{i=1}^{N}, \mathbb{P}, \{R_i\}_{i=1}^{N}, \gamma \rangle$, including a universal state space $\mathcal{S}$, ``perspectives" $ \{\mathcal{S}_i\}_{i=1}^N$, joint action space $ \mathcal{A} = \mathcal{A}_1\times\cdots\times\mathcal{A}_N$, transition dynamics $\mathbb{P}:\mathcal{S}\times\mathcal{A}\rightarrow\Delta(\mathcal{S})$, reward functions $R_i:\mathcal{S}\times\mathcal{A}\times\mathcal{S}\rightarrow\mathbb{R}$, and discount $\gamma$. The goal is to maximize the expected discounted sum of future reward for each agent $\tilde{V}^{\pi}_i(s_0) = \mathbb{E} \left[ \sum_{t\geq0} \gamma^t R_i\left(s_t,\pi(s_t), s_{t+1}\right) \right]$, where $s_0\in\mathcal{S}$ is an initial universal state and $\pi=(\pi_1,\dots,\pi_N)$ is a joint policy. Similarly:
\begin{equation}\label{eq:q_function_mg}
    \tilde{Q}^{\pi}_i(s_0, a_0) = \mathbb{E} \left[ R_i(s_0, a_0, s_1) + \sum_{t\geq1} \gamma^t R_i\left(s_t, \pi(s_t), s_{t+1} \right) \right].
\end{equation}
Each $\tilde{V}^{\pi}_i$ is meant to be maximized with respect to all other policies, such that
\begin{equation*}
    \pi_i=\underset{\pi_i'}{\rm argmax} \ \tilde{V}^{\pi'}_i \text{ where } \pi'=(\pi_1, \dots, \pi_i',\dots, \pi_N).
\end{equation*}
IQL can be employed to optimize each $\pi_i$. Moreover, if $\{\mathcal{S}_i\}_{i=1}^{N}$, $\{\mathcal{A}_i\}_{i=1}^{N}$, and $\{R_i\}_{i=1}^{N}$ are similar among all agents, parameter sharing can be employed to learn the policies.

\subsection{Police Operations}
Police jurisdictions are typically hierarchically divided. For example, the Atlanta Police Department (APD) divides the city of Atlanta into six \textit{zones}, each of which is divided into \textit{beats} \citep{apdzones}. One patroller is usually assigned to each beat. When calls-for-service arrive at the dispatcher, an officer in the same zone as the incident is dispatched to the scene; a patroller can be dispatched to a call outside their assigned beat. This joint problem of beat-based patrol and dispatch is further discussed and formalized in Section~\ref{sec:formulation}.

\section{Problem Formulation}\label{sec:formulation}
Consider patrollers $i=1,\dots,N$ making decisions in time $t=1,2,\dots$. Joint patrol and dispatch is formalized as an MDP with a central agent and then is decomposed into an MG of $N+1$ agents.

\begin{figure}[t]
    \centering
    \includegraphics[width=0.5\textwidth]{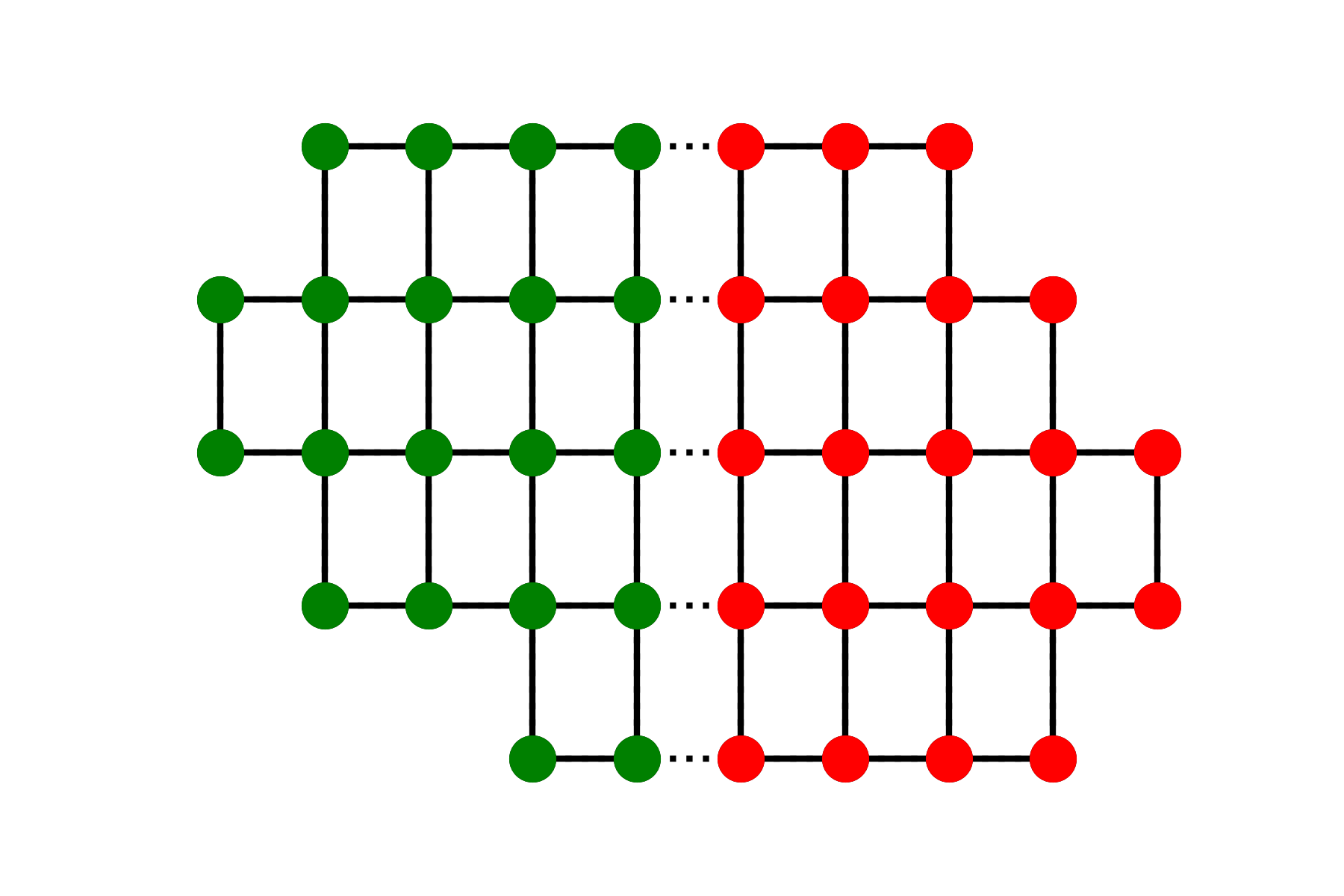}
    \caption{Example of a graph with two beats indicated by color. The dotted edges indicate that patrollers can travel between beats only when responding to calls.}
    \label{fig:graph_example}
\end{figure}

\subsection{The Patrol Problem}\label{sec:patrol_problem}
Patrollers navigate a connected, undirected graph $G=(V,E)$ \citep{almeida2004recent} with nodes/locations $V$ connected by edges $E$ which are traversed in a single iteration; each $v\in V$ is connected to itself. Each patroller $i$ is assigned to a node subset (beat) $V_i$, an example is visualized in Figure~\ref{fig:graph_example}. The \textit{patrol problem} involves travel decisions for each patroller. For patrollers not busy responding to incidents (Sections~\ref{sec:dispatch_problem} and \ref{sec:patrol_and_dispatch}), a \textit{patrol policy} directs each patroller to a neighboring node in $V_i$.

\subsection{The Dispatch Problem}\label{sec:dispatch_problem}
Incidents of categories $k=1,\dots,K$ arrive with rates $\lambda_k$ following distributions $q_k(v)$ supported on $V$, requiring $t_{\rm scene}\sim{\rm Exp}(\beta_k)$ on-scene time to manage. The \textit{dispatch problem} involves matching \textit{free patrollers} (not currently dispatched) to \textit{idle incidents} (not yet been assigned). A \textit{dispatch policy} selects among all possible groups of pairings, including leaving incidents idle or patrollers free. A dispatched patroller travels along the shortest path towards the incident, whose number of edges is equivalent to the travel time $t_{\rm travel}$ in iterations. If a patroller arrives at iteration $t$, it is finished managing the incident starting at iteration $t+t_{\rm scene}$.

\subsection{Joint Patrol \& Dispatch}\label{sec:patrol_and_dispatch}
The service time for patroller $i$ to manage incident $j$ is $t_{\rm service}(i,j) = t_{\rm idle}(j) + t_{\rm travel}(i, j) + t_{\rm scene}(j)$, where $t_{\rm idle}(j)$ is the time spent in the queue before dispatch. When patroller $i$ is dispatched, it must travel toward the incident or remain at the scene. When outside its beat, it must return once free. Define the response time, reflecting the time incident $j$ waited before patroller $i$ arrived:
\begin{equation}\label{eq:response_time}
    t_{\rm response}(i,j) = t_{\rm idle}(j) + t_{\rm travel}(i, j).
\end{equation}
Joint patrol and dispatch can be formulated as an MDP to learn joint policy $\pi$.

\subsubsection{Joint State Space.}\label{sec:joint_state_space}
Given patroller positions $v_1,\dots,v_N\in V$ and busy statuses $u_1,\dots,u_N\in\mathbb{N}$ (travel and on-scene time), define ${\bf v} := \left[(v_1,u_1),\dots,(v_N,u_N)\right]$. At dispatch, patrollers observe the random on-scene service time. Given locations of $M$ idle incidents $w_1,\dots,w_M\in V$ and idle times $p_1,\dots,p_M\in\mathbb{N}$, define ${\bf w} := \left[(w_1,p_1),\dots,(w_M,p_M)\right]$. The queue has a fixed idle-incident capacity $M$; when full and a new incident arrives, the longest-waiting incident is replaced. The joint state space is $\mathcal{S} = \left\{s = [{\bf v} \oplus {\bf w}]^T\right\}$, where $\oplus$ refers to vector concatenation.

\subsubsection{Joint Action Space.}\label{sec:joint_action_space}
For a free patroller, patrol action set $\mathcal{A}^{(\rm patrol)}_i(v_{i,t}, u_{i,t})$ contains movements to neighboring vertices in $V_i$. A busy patroller moves towards an incident or stays at the scene, moving back towards its beats afterward. After patrol actions, a dispatch action is selected from the dispatch action set $\mathcal{A}^{(\rm dispatch)}({\bf v}'_t, {\bf w}'_t)$. Note, ${\bf v}'_t$ and ${\bf w}'_t$ refer to the updated ${\bf v}_t$ and ${\bf w}_t$ according to the patrol actions and the incident arrivals. The joint action space is
\begin{equation*}
    \mathcal{A}({\bf v}_t, {\bf w}_t) = \mathcal{A}^{\rm (patrol)}_1(v_{1,t},u_{1,t})\times\cdots\times\mathcal{A}^{\rm (patrol)}_N(v_{N,t},u_{N,t})\times\mathcal{A}^{(\rm dispatch)}({\bf v}'_t, {\bf w}'_t).
\end{equation*}
The patrol action space has exponential cardinality in $N$, and the dispatch action space has combinatorial cardinality.

\subsubsection{Response-Based Reward \& Objective.}\label{sec:reward_function}
The reward function is the negative sum of response times and a penalty for incidents removed from a capacitated queue:
\begin{equation}\label{eq:rl_reward}
    R(t) = - \left( \sum_{j \text{ dispatched in iteration } t} t_{\rm response}(i, j) + \sum_{j \text{ removed in iteration } t} \alpha\cdot t_{\rm idle}(j)  \right).
\end{equation}
Penalty weight $\alpha$ is a hyper-parameter ($\alpha=2$ in this work). Given $s_0\in\calS$,
\begin{equation*}
    \pi^\star (s_0) = \underset{\pi\in\Pi}{\rm argmax} \ \mathbb{E} \left[ \sum_{t\geq0} \gamma^t R(t) \ \middle| \ s_t \sim \mathbb{P} \left(s_{t-1},\pi(s_{t-1})\right) \forall t>0 \right].
\end{equation*}

\subsubsection{Markov Game Decomposition.}\label{sec:mg_decomp}
Patrol and dispatch can be decomposed into an MG with $N+1$ agents defined by tuple:
\begin{equation*}
    \langle N+1, \ \mathcal{S}, \ \left\{\mathcal{S}^{\rm(patrol)}_i\right\}_{i=1}^{N}\cup\mathcal{S}^{\rm(dispatch)}, \ \left\{\mathcal{A}^{\rm(patrol)}_i\right\}_{i=1}^{N}\cup\mathcal{A}^{\rm(dispatch)}, \ \mathbb{P}, \ R, \ \gamma \rangle.    
\end{equation*}
Patrol agent $i$ corresponds to $\mathcal{A}^{\rm(patrol)}_i(v_{i,t},u_{i,t})$ and the dispatch agent corresponds to $\mathcal{A}^{(\rm dispatch)}({\bf v}'_t, {\bf w}'_t)$. Similar but distinct state spaces $\{\mathcal{S}^{\rm(patrol)}_i\}_{i=1}^N$ are defined so that each patrol agent observes its own perspective, swapping the location and busy status to the first index: ${\bf v}^{(i)}:= \left[(v_i,u_i),\dots,(v_{i-1},u_{i-1}),(v_1,u_1),(v_{i+1},u_{i+1}),\dots,(v_N,u_N)\right]$. This yields $\mathcal{S}^{\rm(patrol)}_i = \left\{s = [{\bf v}^{(i)} \oplus {\bf w}]^T\right\}$. The dispatch agent observes joint state space $\mathcal{S}^{\rm(dispatch)} = \mathcal{S}$. Reward $R(t)$ \eqref{eq:rl_reward} is shared. The goal is to learn patrol policies $\pi^{\rm(patrol)}_i$ and dispatch policy $\pi^{\rm(dispatch)}$ to minimize the response time while avoiding incidents being removed from the queue.

\begin{figure}[t]
    \centering
    \includegraphics[width=0.75\textwidth]{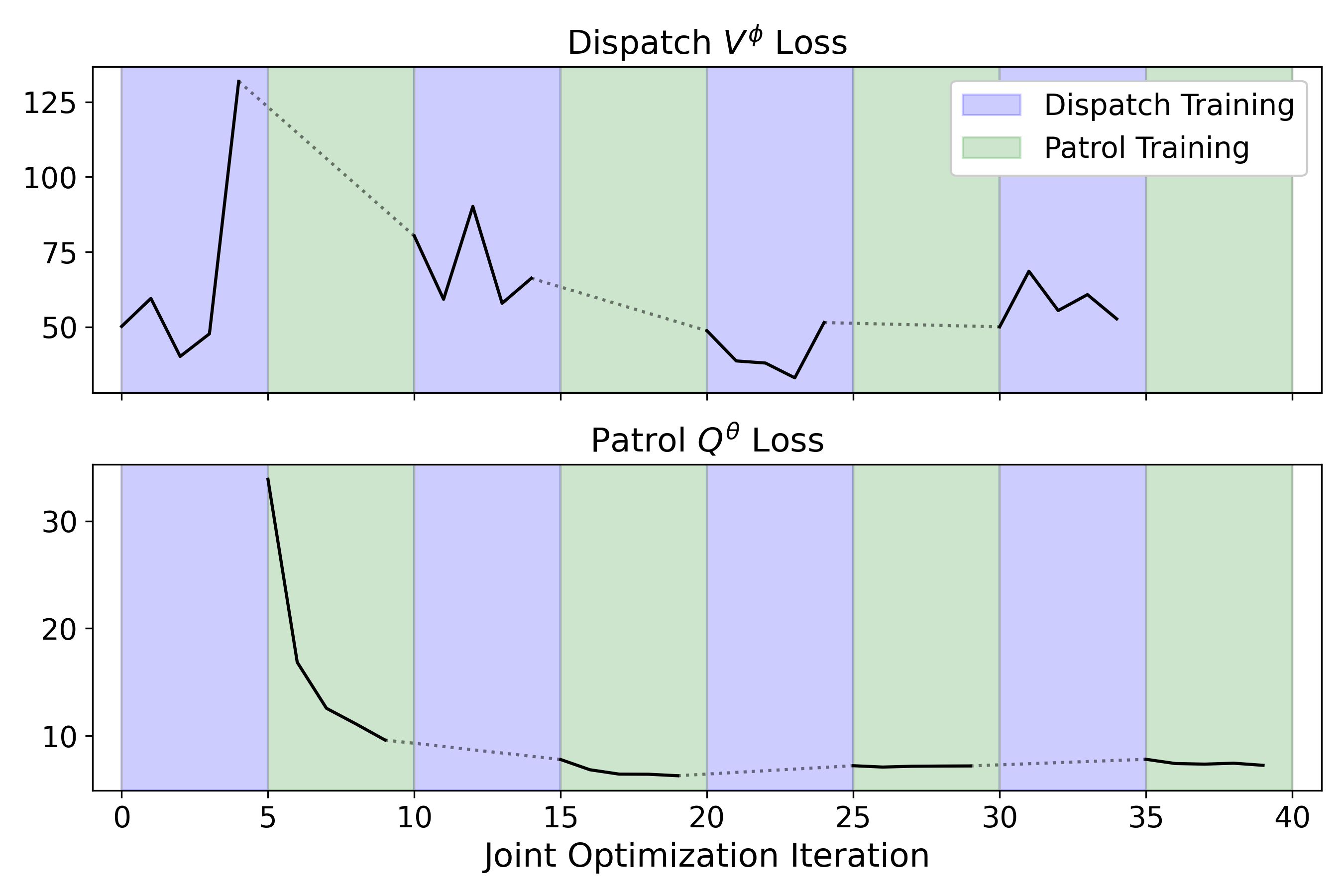}
    \caption{Loss of the dispatch and patrol value networks over the course of alternating joint optimization in the high call volume setting (Section~\ref{sec:efficient_response_experiment}).}
    \label{fig:sim_loss_disp_pat}
\end{figure}

\section{Methodology}\label{sec:method}
Parameter sharing in deep $Q$-learning is applied to learn patrol policies for each patroller (Section~\ref{sec:marl_patrolling}), and incidents are assigned to patrollers using a MIP and value function approximation (Section~\ref{sec:dispatch_mip}). These are combined to learn joint policies for patrol and dispatch (Section~\ref{sec:joint_patrol_dispatch_policy}).

\subsection{MARL with Parameter Sharing for Patrolling}\label{sec:marl_patrolling}
The MG patrol policies (Section~\ref{sec:mg_decomp}) are learned using MARL. {Recall, in Section~\ref{sec:formulation}, the state spaces $\{\mathcal{S}^{\rm(patrol)}_i\}_{i=1}^N$ are defined to contain the positions and statuses of all patrollers and incidents on the graph, while the action spaces $\{\mathcal{A}^{\rm(patrol)}_i\}_{i=1}^N$ contain movements along the graph edges. Furthermore, all patrol agents share a reward $R$. Based upon prior MARL works indicating that similar agents benefit from parameter sharing~\citep{gupta2017cooperative}, this similarity in the state spaces, action spaces, and shared reward suggests that parameter sharing can be applied.}

{Moreover, since the state spaces and action spaces are nearly identical, we employ an extreme version of parameter sharing in learning the patrol policies. That is,} a single neural network $Q^\theta$ (parameterized by $\theta$) learns a state-action value function for all agents. {Our assumption is that the universal approximation power of a ``large enough'' neural network has the capacity to jointly learn all patrol policies, which is validated by the performance of the learned patrol policies in the experiments of Section~\ref{sec:experiments}. We find that DQNs of relatively few layers can learn these policies. Patrol policies thus select actions according to $\pi_i^\theta(s_i) = \underset{a}{\rm argmax} \  Q^\theta(s_i,a)$.} Mini-batches of transitions $(s_{i,t}, a_{i,t}, r_t, s_{i,t+1})$ from each agent's perspective are collected via $\epsilon$-greedy simulation to update $Q^\theta$ using LSVI updates.

\subsection{Policy Iteration for Dispatching}\label{sec:dispatch_mip}
A variation of the MIP action selection strategy outlined in Section~\ref{sec:combinatorial_rl} is employed for dispatch. Let $x_{ij}=1$ if patroller $i$ is assigned to incident $j$, otherwise let $x_{ij}=0$. The first term of reward \eqref{eq:rl_reward} can be computed given $x_{ij}$: $\sum_{i,j} t_{\rm response}(i,j)\cdot x_{ij}$. Unlike \cite{delarue2020reinforcement}, we cannot directly compute the entire reward, nor can we obtain the transition state deterministically due to incident queue ``overflow'' resulting from stochastic call arrival dynamics. This factor, and the difference in patroller positions due to the patrol policy, affects the state transition.

Given $V^{\rm(dispatch)}$ (or an approximation $V^\phi$), the next-state value can be approximated. Define the \textit{value delta} for patroller $i$ as
\begin{equation}\label{eq:patrol_delta}
    \delta V^{\rm(dispatch)}_i(s) := \mathbb{E} \left[ V^{\rm(dispatch)}(\text{patroller } i \text{ assigned}) - V^{\rm(dispatch)}(\text{patroller } i \text{ not assigned}) \right]
\end{equation}
and for incident $j$ as
\begin{equation}\label{eq:incident_delta}
    \delta V^{\rm(dispatch)}_j(s) := \mathbb{E} \left[ V^{\rm(dispatch)}(\text{incident } j \text{ assigned}) - V^{\rm(dispatch)}(\text{incident } j \text{ not assigned}) \right],
\end{equation}
where the expectation is taken under the transition dynamics $\mathbb{P}$ conditioned upon the policies for patrol and dispatch. These represent beginning in state $s$ and considering the expected change in value for assigning patroller $i$ to some idle incident, with no other assignments made. Given approximators $\delta V^{\phi_1}_i$ for $\delta V^{\rm(dispatch)}_i$ and $\delta V^{\phi_2}_j$ for $\delta V^{\rm(dispatch)}_j$, dispatch is formulated as the MIP:
\begin{equation*}
    \begin{aligned}
        &\pi^{\phi}(s) =  & \underset{x}{\text{argmin}}\quad & \sum_{i=1}^N \sum_{j=1}^M t_{\rm response}(i,j)\cdot \left(x_{ij} - \delta V^{\phi_1}_i(s) - \delta V^{\phi_2}_j(s)\right)\\
        & &\text{s.t.}\quad & \sum_{j=1}^M x_{ij} \leq \mathbb{I}_{\{\text{patroller } i \text{ is free}\}} \quad \forall i\in\{1,\dots,N\} \\
        & & & \sum_{i=1}^N x_{ij} \leq \mathbb{I}_{\{\text{incident } j \text{ is idle}\}} \quad \forall j\in\{1,\dots,M\} \\
        & & & x_{ij}\in\{0,1\} \quad \forall i\in\{1,\dots,N\},j\in\{1,\dots,M\},
    \end{aligned}
\end{equation*}
where $\mathbb{I}_{\{\text{event}\}}$ is an indicator for $\{\text{event}\}$. Notably, given $\delta V^{\phi_1}_i$ and $\delta V^{\phi_2}_j$, this can be computed deterministically given $x_{ij}$ and the initial state $s$.

$\delta V^{\phi_1}_i$ and $\delta V^{\phi_2}_j$ can be learned by first approximating $V^{\rm(dispatch)}$ using $V^\phi$ (Section~\ref{sec:value_learning}). Using $V^\phi$, mappings are generated from states to expectations \eqref{eq:patrol_delta}-\eqref{eq:incident_delta}. Approximators $\delta V^{\phi_1}_i$ and $\delta V^{\phi_2}_j$ are trained to map states to value deltas. We utilize a policy iteration scheme as in \cite{delarue2020reinforcement}; we iterate between updating $V^\phi$ and updating $\delta V^{\phi_1}_i$ and $\delta V^{\phi_2}_j$ using $V^\phi$. In practice, one network is learned for patrollers, $\delta V^{\phi_1}$, and one network for incidents, $\delta V^{\phi_2}$, which have output dimension equal to the number of patrollers and the size of the incident queue, respectively.

\subsection{Joint Patrol \& Dispatch Policy }\label{sec:joint_patrol_dispatch_policy}
{The joint policy for patrol and dispatch combines the separate patrol and dispatch policies by optimizing their parameters alternatively. As described in Section~\ref{sec:joint_opt}, the parameters of the patrol policies are frozen when learning those of the dispatch policy, and vice versa. This alternating procedure is empirically found to converge to effective policies that outperform those optimized for patrol or dispatch alone (see Figure~\ref{fig:sim_loss_disp_pat} and Section~\ref{sec:experiments}).}

{Given jointly-optimized policies for patrol and dispatch, the policy is deployed as follows:} patroller $i$ observes a state and moves according to $\pi_i^\theta$. Incidents are sampled, and the dispatch agent matches patrollers to incidents according to $\pi^\phi$. $R(t)$ and joint state $s_{t+1}$ are finally computed. The dispatcher observes the state as influenced by the patrol policy in the first phase, with reward and subsequent state observed by the patrollers being influenced by the dispatch policy. {The assumption that the coordination of the patrollers and the dispatcher is enhanced by a joint optimization is validated by the improved performance of the jointly-optimized policies in our experiments.}

\begin{table}[!t]
  \caption{Response time distribution characteristics (over 100 simulations of 5000 iterations) for the settings of Section~\ref{sec:efficient_response_experiment}. Quantities in parenthesis are standard deviations. Notation $q=Q$ refers to the $Q$-quantile.}
  \centering
  \resizebox{\textwidth}{!}
  {
  \begin{threeparttable}
    \begin{tabular}{lcccccccc}
    \toprule
    \toprule
    & \multicolumn{4}{c}{High Call Volume} & \multicolumn{4}{c}{Low Call Volume} \\
     \cmidrule(lr){2-5} \cmidrule(lr){6-9}
    \multirowcell{2}{\\Policy} & \multirowcell{2}{Avg.\\Response} & \multirowcell{2}{Avg.\\Overflows} & \multirowcell{2}{$q=0.75$\\Response} & \multirowcell{2}{$q=0.95$\\Response} & \multirowcell{2}{Avg.\\Response} & \multirowcell{2}{Avg.\\Overflows} & \multirowcell{2}{$q=0.75$\\Response} & \multirowcell{2}{$q=0.95$\\Response} \\
    \\
    \midrule
    Heuristic& 10.0$_{(5.90)}$ & 131$_{18.6}$ & 14 & 21 & 7.07$_{(4.81)}$ & 6.63$_{(3.82)}$ & 9 & 16  \\
    Patrol    & 9.57$_{(5.83)}$ & 124$_{20.4}$ & 13 & 20 & 6.24$_{(4.47)}$ & 4.61$_{(3.26)}$ & \textbf{8} & 15  \\
    Dispatch& 8.13$_{(5.43)}$ & \textbf{77.5}$_{13.8}$ & \textbf{11} & 19 & 7.16$_{(4.10)}$ & 53.5$_{(10.0)}$ & 10 & 15  \\
    Joint     & \textbf{8.09}$_{(5.33)}$ & 79.8$_{14.6}$ & \textbf{11} & \textbf{18} & \textbf{5.95}$_{(4.18)}$ & \textbf{3.69}$_{(2.69)}$ & \textbf{8} & \textbf{14} \\
    \bottomrule
    \bottomrule
  \end{tabular}
  \end{threeparttable}
  }
  \label{tab:sim_response_time}
\end{table}

\subsubsection{Joint Policy Optimization}\label{sec:joint_opt}
We alternatively optimize the parameters defining the patrol and dispatch policies. One phase involves $Q$-learning of the shared patroller $Q^\theta$, while the other phase involves updates to the incident value networks $V^\phi$, $\delta V_i^{\phi_1}$, and $\delta V_j^{\phi_2}$, repeating to convergence. For outer loop iterations, a number of inner loop updates are made to the dispatcher, followed by a number of inner loop updates to the patrollers. The steps of the $n_{{\rm inner},\phi}$ inner loop updates to the dispatcher are: 
\begin{itemize}
    \item[(1)] collect transitions from on-policy simulation,
    \item[(2)] conduct $n_{{\rm epoch},\phi}$ epochs of training for $V^\phi$,
    \item[(3)] generate ``delta transitions'' for each delta network using $V^\phi$, and 
    \item[(4)] conduct $n_{{\rm epoch},\phi_1}=n_{{\rm epoch},\phi_2}$ epochs of training on each network $\delta V^{\phi_1}$ and $\delta V^{\phi_2}$.
\end{itemize}
The steps of the $n_{{\rm inner},\theta}$ inner loop updates to the patrollers are:
\begin{itemize}
    \item[(1)] collect transitions from $\epsilon$-greedy on-policy simulation and 
    \item[(2)] conduct $n_{{\rm epoch},\theta}$ epochs of training for $Q^\theta$.
\end{itemize}
This is repeated for $n_{{\rm outer}}$ outer loop iterations. We find that performance gains from learning dispatch policies are greater than patrol policies, so we begin optimization with a ``warm-start'' to the dispatch policy consisting of $n_{\rm warm}$ dispatch inner loop iterations. For the experiments of Section~\ref{sec:experiments}, all networks are multi-layer perceptrons (MLPs) using rectified linear unit (ReLU) activation. Dispatch inner loops train using $n_{\rm dispatch}$ number of transitions (per inner loop iteration), while patrol inner loops train using $n_{\rm patrol}$ transitions. All transitions are split into an 80\%/20\% train/validation split. The discount factor $\gamma=0.9$ in all settings.

\section{Experiments}\label{sec:experiments}
The experiments compare joint optimization of patrol and dispatch to policies optimized for patrolling or dispatch alone. In Section~\ref{sec:efficient_response_experiment}, the goal is to minimize the reward as defined in \eqref{eq:rl_reward}. Then, in Section~\ref{sec:equity_experiment}, the reward is altered to demonstrate learning of more equitable policies. Finally, a real data example is outlined in Section~\ref{sec:apd_cfs} based on a region in the City of Atlanta. See Appendix~\ref{app:model_selection} for a discussion of model selection in each experiment.

\paragraph{Baselines \& Comparison.} 
The three baselines include patrol-only optimization, dispatch-only optimization, and a fully heuristic policy. For policies without optimized patrol, patrollers move randomly in their assigned regions. For policies without optimized dispatch, dispatch follows a first come, first serve priority queue whereby incident class 2 precedes incident class 1, and the closest free patroller is sent to the scene. To compare policies, we evaluate incident response over $n_{\rm episode}$ episodes, each consisting of 5,000 iterations of on-policy simulation. The characteristics of the response time distributions are compared in addition to the number of incident queue overflows per episode.

\begin{figure}[t]
    \centering
    \includegraphics[width=\textwidth]{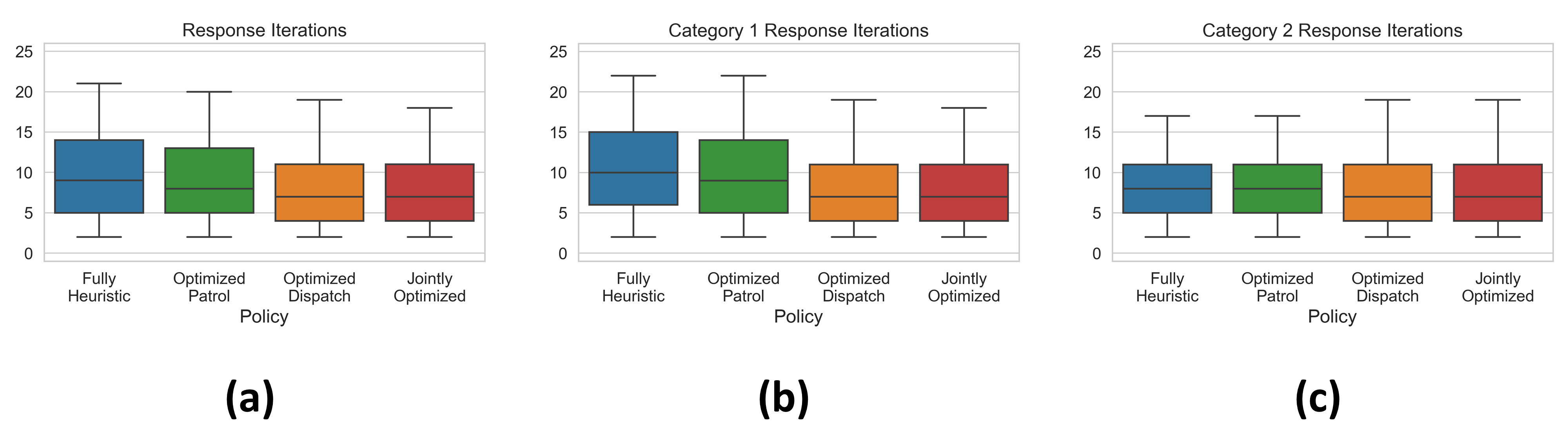}
    \caption{Response time distributions (a) for all incidents and (b)-(c) for each incident category in the high call volume setting of Section~\ref{sec:efficient_response_experiment}.}
    \label{fig:high_vol_response_times_violin}
\end{figure}

\begin{figure}[t]
    \centering
    \includegraphics[width=\textwidth]{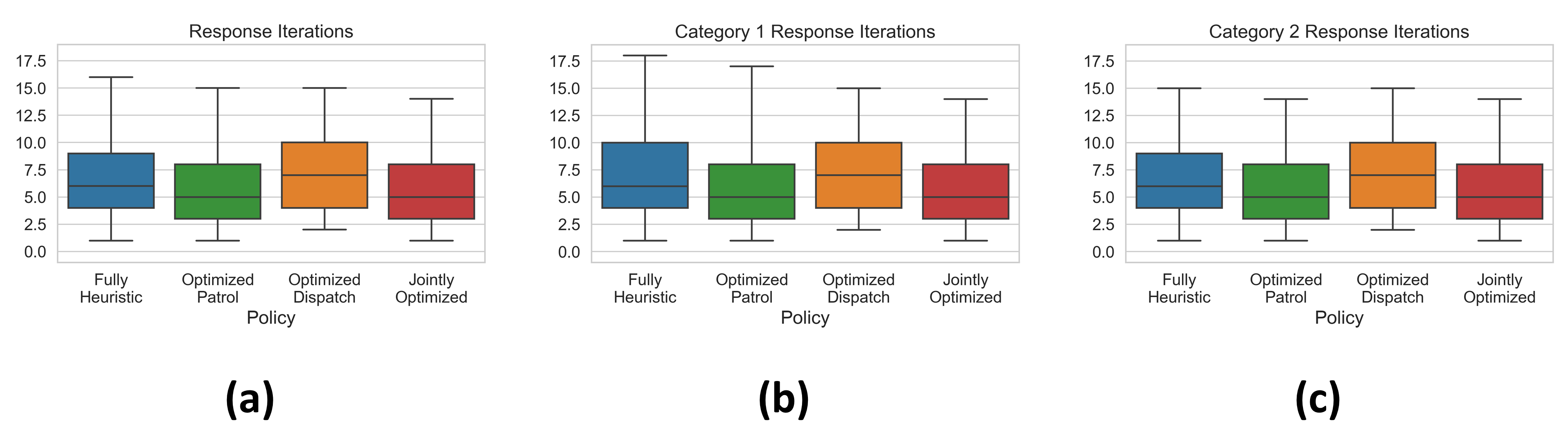}
    \caption{Response time distributions (a) for all incidents and (b)-(c) for each incident category in the low call volume setting of Section~\ref{sec:efficient_response_experiment}.}
    \label{fig:low_vol_response_times_violin}
\end{figure}

\subsection{Efficient Response Experiments}\label{sec:efficient_response_experiment}

\paragraph{Simulated Settings.}
Consider two simulated settings on the same graph consisting of two $7\times7$ beats adjacent on one edge. Two categories of incident arrive with rates $\lambda_1$ and $\lambda_2$, uniform distributions $q_1$ and $q_2$, and average service times $\beta_1=1$ and $\beta_2=3$. The incident queue size is 3. In the \textbf{High Call Volume} setting, $\lambda_1=0.15$ and $\lambda_2=0.075$. In the \textbf{Low Call Volume} setting, $\lambda_1=0.075$ and $\lambda_2=0.05$.

\paragraph{Modeling \& Optimization.}
Joint optimization is conducted using $n_{\rm outer}=4$ with $n_{{\rm inner},\phi}=5$ and $n_{{\rm inner},\theta}=5$ and beginning with $n_{\rm warm}=20$. Each dispatch inner loop trains for $n_{{\rm epochs},\phi}=n_{{\rm epochs},\phi_1}=n_{{\rm epochs},\phi_2}=25$ epochs each for $V^\phi$, $\delta V^{\phi_1}$, and $\delta V^{\phi_2}$. This is conducted over $n_{\rm dispatch}=1,000$ simulated transitions per iteration with batch size 100 and learning rate $10^{-3}$. Each patrol inner loop fixes $\epsilon=1$ (fully random patrol) and trains for $n_{{\rm epochs},\theta}=1$ epochs for $Q^\theta$, which is conducted over $n_{\rm patrol}=1.25$ million simulated transitions with batch size 50 and learning rate $10^{-5}$. $V^\phi$, $\delta V^{\phi_1}$, and $\delta V^{\phi_2}$ are MLPs of 1 hidden layer with 128 nodes, while $Q^\theta$ is an MLP of 2 hidden layers with 512 nodes each. Validation loss is visualized in Figure~\ref{fig:sim_loss_disp_pat} for the high call volume setting. The dispatch dynamic is very noisy but still leads to an effective policy. The patrol-only optimization uses $n_{{\rm inner},\theta}=20$ updates to a patrol $Q^\theta$ network. The dispatch-only optimization uses $n_{{\rm inner},\phi}=50$ updates to dispatch networks. All other patrol- and dispatch-only optimization parameters are identical to those of the joint optimization.

\paragraph{Results.}
Response time distributions are generated using $n_{\rm episodes}=100$. The high call volume experiment results are visualized in Table~\ref{tab:sim_response_time} and Figure~\ref{fig:high_vol_response_times_violin}. The jointly-optimized policy has the fastest response and loses few incidents. This and the dispatch-only optimized policy have similar response distributions. Response times are visualized by category in Figure~\ref{fig:high_vol_response_times_violin}(b)-(c); the jointly-optimized policy is best in category 1 with diminished performance in category 2. This could indicate that dispatch optimization yields policies that prefer incidents that take less time to manage (since $\beta_1<\beta_2$), resulting in lower average response times. Table~\ref{tab:sim_response_time} and Figure~\ref{fig:low_vol_response_times_violin} show the low call volume results. The jointly-optimized policy responds more quickly, and all policies lose a similarly small number of calls except for the dispatch-only optimized policy. In this case, the jointly-optimized policy has a similar response time to both categories.

\begin{figure}[t]
    \centering
    \includegraphics[width=\textwidth]{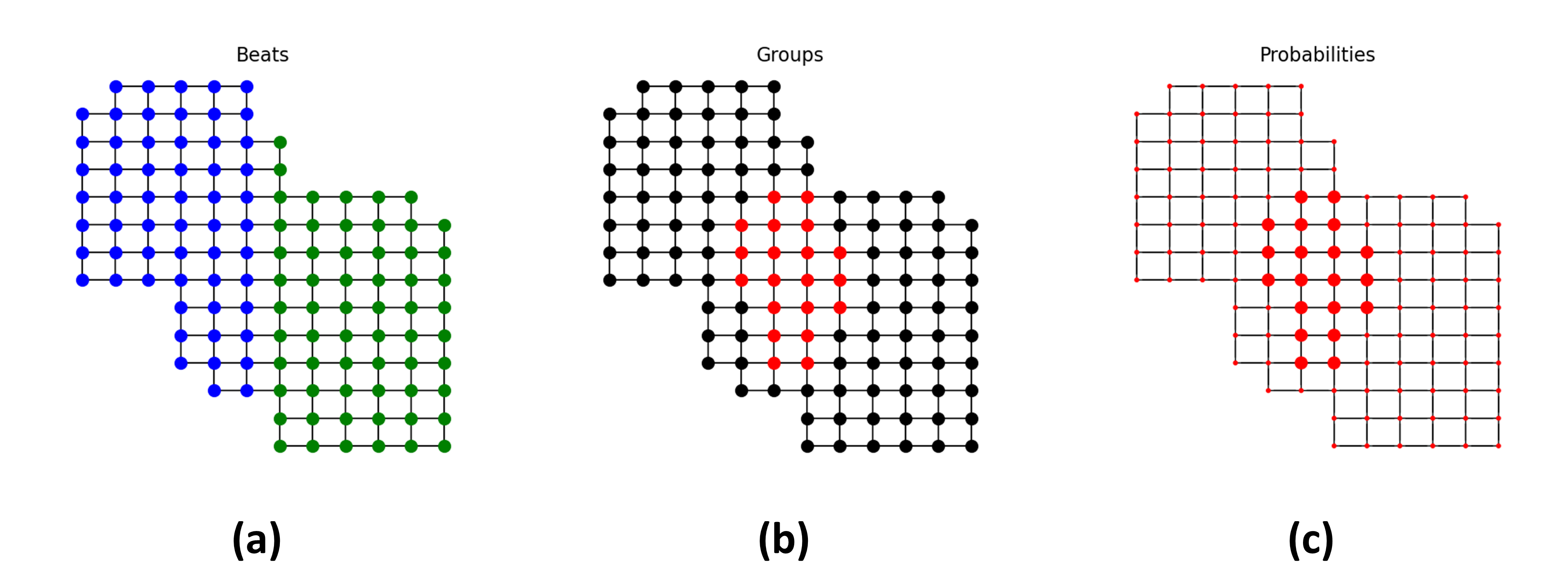}
    \caption{The graph setting of Section~\ref{sec:equity_experiment}, which is divided into two beats in (a), as well as two \textit{groups} in (b) and (c) which experience different rates of incident arrival.}
    \label{fig:fairness_graph}
\end{figure}

\begin{table}[t]
  \caption{Response time distribution characteristics for Section~\ref{sec:equity_experiment}. ``Group Difference'' refers to the difference in average response time between Group 2 and Group 1, and ``Coverage Ratio'' refers to the ratio of the number of iterations spent in Group 2 to Group 1.}
  \centering
  \resizebox{\textwidth}{!}
  {
  \begin{threeparttable}
    \begin{tabular}{lcccccccccccc}
    \toprule
    \toprule
    & \multicolumn{4}{c}{Group 1 (Small)} & \multicolumn{4}{c}{Group 2 (Large)} & \multicolumn{3}{c}{Overall} \\
     \cmidrule(lr){2-5} \cmidrule(lr){6-9} \cmidrule(lr){10-12}
    \multirowcell{2}{\\Policy} & \multirowcell{2}{Avg.\\Response} & \multirowcell{2}{Avg.\\Overflows} & \multirowcell{2}{$q=0.75$\\Response} & \multirowcell{2}{$q=0.95$\\Response} & \multirowcell{2}{Avg.\\Response} & \multirowcell{2}{Avg.\\Overflows} & \multirowcell{2}{$q=0.75$\\Response} & \multirowcell{2}{$q=0.95$\\Response} & \multirowcell{2}{Avg.\\Response} & \multirowcell{2}{Group\\Difference} & \multirowcell{2}{Coverage\\Ratio} \\
    \\
    \midrule
    Heuristic & 9.07$_{(5.10)}$ & 63.5$_{(11.3)}$ & 12 & 19 & 12.0$_{(5.66)}$ & 30.8$_{(7.03)}$ & 16 & \textbf{22} & 10.0$_{(5.47)}$ & 2.96 & 0.91 \\
    
    $\rho_{\rm large}=0.5$      & 8.95$_{(6.16)}$   & 63.4$_{(11.7)}$  & 11 & 21  & \textbf{11.7}$_{(6.20)}$   & \textbf{29.9}$_{(6.21)}$ & \textbf{15} & 24 & 9.86$_{(6.31)}$ & \textbf{2.75} & \textbf{1.40} \\
    
    $\rho_{\rm large}=1.0$      & \textbf{8.11}$_{(5.46)}$   & \textbf{49.3}$_{(10.8)}$  & \textbf{10} & \textbf{18}  & 11.8$_{(7.50)}$   & 38.8$_{(8.37)}$ & \textbf{15} & 26 & \textbf{9.28}$_{(6.42)}$ & 3.66 & 0.93 \\
    
    \bottomrule
    \bottomrule
  \end{tabular}
  \end{threeparttable}
  }
  \label{tab:fair_response_time}
\end{table}

\subsection{Equity-based Reward}\label{sec:equity_experiment}

\paragraph{Simulated Setting.}
Figure~\ref{fig:fairness_graph}(a) visualizes a graph of 119 nodes divided into two beats of 58 and 61 nodes. One incident category arrives with rate $\lambda=0.2$ and has average service time $\beta=3$. In Figure~\ref{fig:fairness_graph}(b) and (c), it is revealed that the non-uniform incident distribution induces two \textit{groups}. The smaller of these groups, \textbf{Group 1}, consists of 20 nodes and has incident probability ten times that of the larger group, \textbf{Group 2}, which consists of 99 nodes.

\paragraph{Fairness.}
An equity-based reward can be constructed to yield different penalties for each group:
\begin{equation}\label{eq:eq_reward}
    \tilde{R}(t) = - \left( \sum_{j \text{ dispatched in iteration } t} \rho_j t_{\rm response}(i, j) + \sum_{j \text{ removed in iteration } t} \rho_j\alpha\cdot t_{\rm idle}(j)  \right),
\end{equation}
where $\rho_j$ is the weight of the penalty incurred for the group in which $j$ occurred. Additional metrics can be introduced to quantify fairness. The distribution of response times can be compared between the two groups; connecting to the independent notion of fairness \citep{barocas2017fairness}, response efficiency should be independent of the group. Similarly, a fair policy could be expected to cover each group for a fraction of iterations equal to the portion of the graph the group represents.

\paragraph{Modeling \& Optimization.}
The networks $V^\phi$, $\delta V^{\phi_1}$, $\delta V^{\phi_2}$, and $Q^\theta$ are structured and optimized exactly as in Section~\ref{sec:efficient_response_experiment}. Jointly-optimized policies trained using equitable reward \eqref{eq:eq_reward} with different weights are compared. Specifically, the weight of the small group is fixed at $\rho_{\rm small}=1$, and the weight of the large group is $\rho_{\rm large}\in\{0.5, 1.0\}$ representing weighted and unweighted policies, respectively.

\begin{figure}[t]
    \centering
    \includegraphics[width=\textwidth]{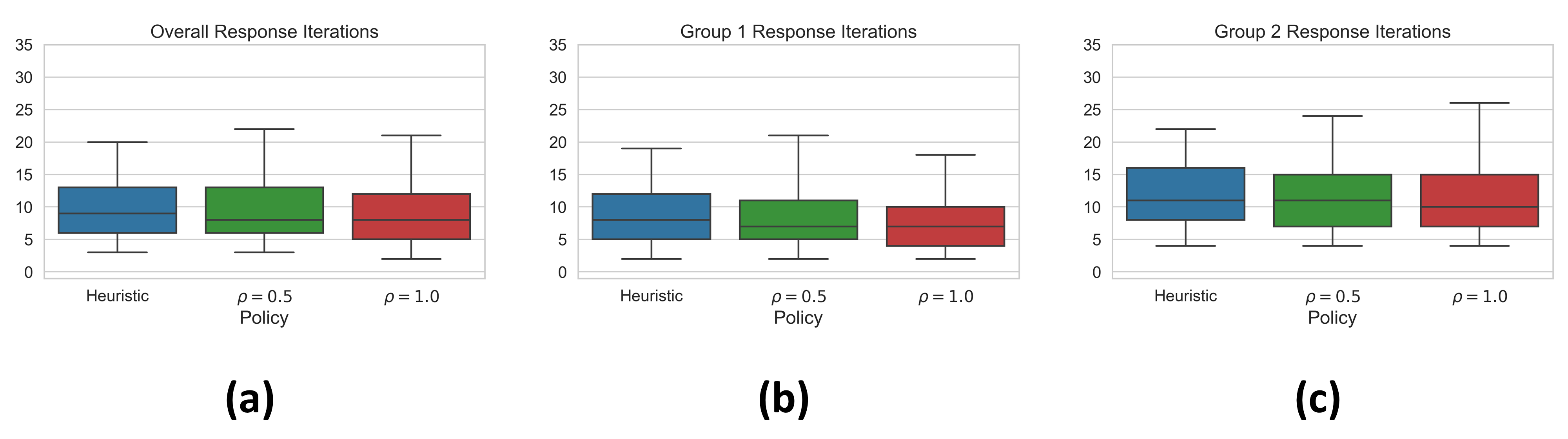}
    \caption{Response time distributions (a) for all groups and (b)-(c) for each group in the equity-based setting of Section~\ref{sec:equity_experiment}.}
    \label{fig:fair_response_time}
\end{figure}

\paragraph{Results.}
Response characteristics are compared using $n_{\rm episodes}=100$. The results are visualized in Table~\ref{tab:fair_response_time} and Figure~\ref{fig:fair_response_time}. A discrepancy in average response time between the two groups is observed in all settings, but this is particularly pronounced in the ``unfair'' optimized setting ($\rho_{\rm large}=1.0$). This discrepancy is about an iteration smaller for the ``fair'' ($\rho_{\rm large}=0.5$), which is also an improvement over the heuristic. Moreover, the fair policy improves upon the overall response time of the heuristic policy while achieving a better balance in response. Coverage is compared using $n_{\rm episodes}=25$ simulations, and the results are compared in the final column of Table~\ref{tab:fair_response_time}. The unfair and heuristic policies do a poor job of balancing coverage, while the fair policy does much better to cover the larger group. Appendix~\ref{app:model_selection} outlines model selection for this section; Figure~\ref{fig:model_selection_section_2} reveals that the unfair ($\rho_{\rm large}=1.0$) policy has a large discrepancy in average response throughout training, which is not the case for the fair policy. Both policies can experience similar coverage ratios, but this can come at a cost to a balanced or effective response.

\begin{figure}[t]
    \centering
    \includegraphics[width=\textwidth]{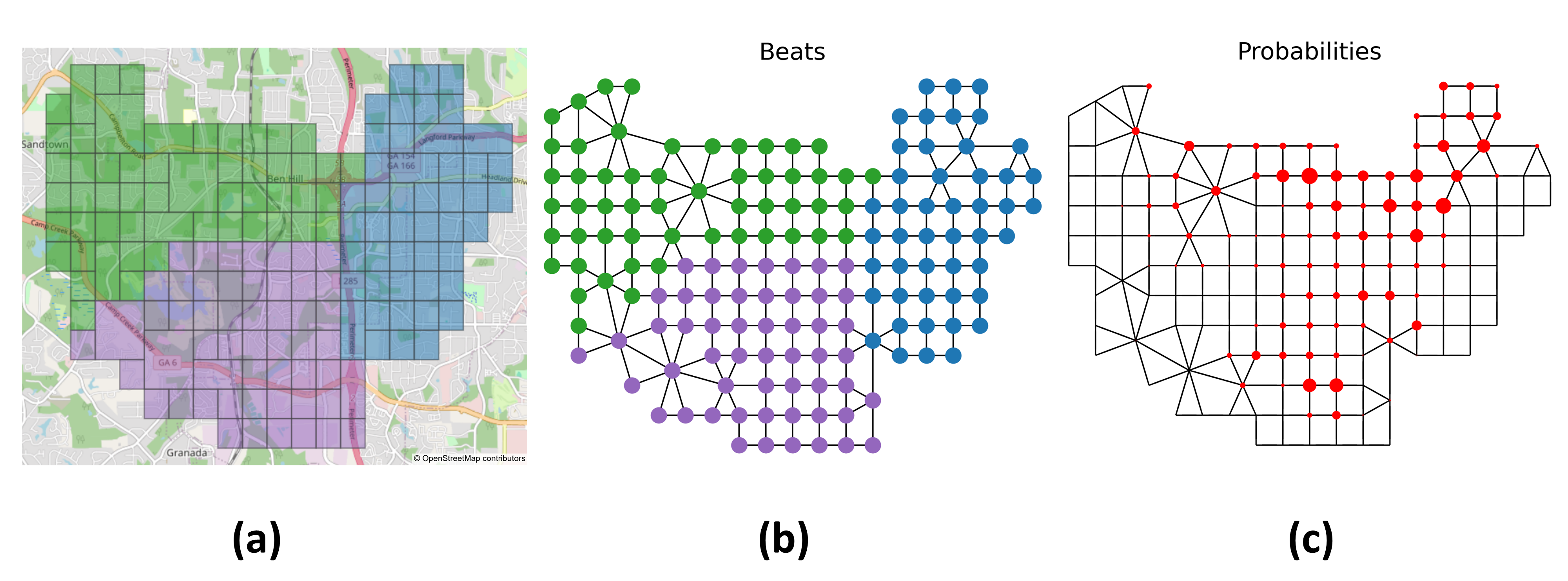}
    \caption{A region of Southwest Atlanta is partitioned into rectangular patches in (a), which are translated into a graph structure composed of 3 beats pictured in (b). In (c), the spatial distribution of incidents estimated from the calls-for-service data is visualized.}
    \label{fig:apd_graph}
\end{figure}

\begin{figure}[t]
    \centering
    \includegraphics[width=\textwidth]{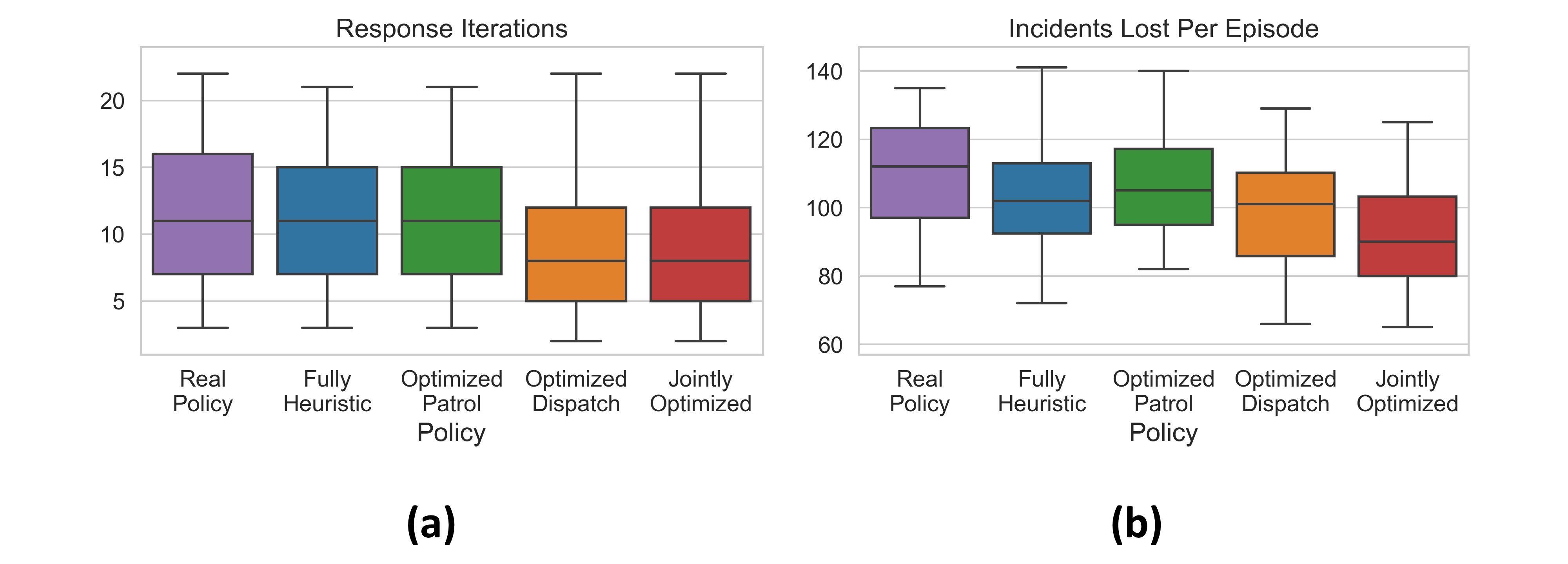}
    \caption{In (a), response time distributions are visualized for the application to the setting based on the city of Atlanta described in Section~\ref{sec:apd_cfs}. Similarly, (b) displays the distributions of incidents lost per episode.}
    \label{fig:apd_response_time}
\end{figure}

\begin{table}[!t]
  \caption{Response time distribution characteristics for the setting based on the real data from the city of Atlanta described in Section~\ref{sec:apd_cfs}.}
  \centering
  \begin{threeparttable}
    \begin{tabular}{lcccc}
    \toprule
    \toprule
    \multirowcell{2}{\\Policy} & \multirowcell{2}{Avg.\\Response} & \multirowcell{2}{Avg.\\Overflows} & \multirowcell{2}{$q=0.75$\\Response} & \multirowcell{2}{$q=0.95$\\Response} \\
    \\
    \midrule
    Real        & 11.7$_{(5.62)}$ &  110$_{17.8}$   & 16 &  22 \\
    Heuristic   & 11.4$_{(5.73)}$ &  104$_{20.5}$   & 15 & \textbf{21}  \\
    Patrol      & 11.3$_{(5.74)}$ &  107$_{18.5}$   & 15 & \textbf{21}  \\
    Dispatch    & 9.50$_{(6.91)}$ &  99$_{19.2}$    & \textbf{12} & 22  \\
    Joint       & \textbf{9.39}$_{(6.66)}$ &  \textbf{92}$_{17.2}$    & \textbf{12} & 22  \\
    \bottomrule
    \bottomrule
  \end{tabular}
  \end{threeparttable}
  \label{tab:apd_response_time}
\end{table}

\subsection{Southwest Atlanta Beats}\label{sec:apd_cfs}

\paragraph{Environment Setting.}
The graph environment in this experiment is based upon a 3-beat region in Southwest Atlanta, which is discretized into 154 rectangular partitions representing nodes on the spatial graph. See Figure~\ref{fig:apd_graph}(a); the patches are selected based on APD patroller trajectory data discretized into 1-minute intervals. The patches are constructed such that observed trajectories take approximately 1 iteration (i.e., 1 minute) to travel between adjacent patches. In Figure~\ref{fig:apd_graph}(b), the resultant graph is shown, which is divided into three beats with one patroller per beat. Finally, as shown in Figure~\ref{fig:apd_graph}(c), incidents are spatially distributed according to a distribution estimated given the $\sim$90k APD calls-for-service from January 1, 2009 to January 8, 2012. The temporal intensity $\lambda=0.25$ per iteration and service time $\beta=5$ iterations.

\paragraph{Real Policy.}
An additional baseline is developed for this setting based on limited APD patroller trajectory data. That is, divided into 1-minute time intervals, APD patroller trajectories over a 2-day interval are tracked along the graph structure. Transitions to adjacent nodes are counted, and the resultant policy ranks potential actions according to the frequency with which each edge was traversed in the real trajectory data. This is denoted the ``Real Policy,'' and can reflect some information about how the police patrol in practice. This policy also uses a first come, first serve queue for dispatch.

\paragraph{Modeling \& Optimization.}
Joint optimization is conducted using $n_{\rm outer}=8$ with $n_{{\rm inner},\phi}=10$ and $n_{{\rm inner},\theta}=5$ with 0 warm-start iterations. The dispatch networks are trained using $n_{\rm dispatch}=10,000$ simulated transitions per inner loop iteration, while all other optimization parameters for patrol and dispatch are identical to those described in Section~\ref{sec:efficient_response_experiment}. For comparison, patrol-only optimization uses $n_{{\rm inner}, \theta}=40$ and dispatch-only optimization uses $n_{{\rm inner},\phi}=80$. As before, the patrol- and dispatch-only optimization parameters are identical to those of the joint optimization.

\paragraph{Results.}
In Figure~\ref{fig:apd_response_time} and Table~\ref{tab:apd_response_time}, the response time distributions for each policy are represented over $n_{\rm episodes}=100$ episodes. While the jointly-optimized policy has the fastest response on average, the overall response time distribution characteristics are similar to those of the dispatch-only optimized policy. However, as shown in Figure~\ref{fig:apd_response_time}(b), the jointly-optimized policy loses substantially fewer incidents per episode than the other policies, indicating that it is the most effective overall policy. Moreover, the jointly-optimized policy leads to faster response and fewer incidents lost than the realistic policy. This suggests that (joint) policy optimization can lead to substantial improvements when applied to real-world environments.

\section{Conclusion}\label{sec:conclusion}
We have outlined a novel procedure for the joint optimization policies for patrol and dispatch using MARL. The patrol policies are learned using deep $Q$-learning with parameter sharing, and the dispatch policies are learned using a policy iteration scheme based on MIP action selection. These policies are combined in a joint optimization procedure that follows a coordinate-descent-style scheme, iteratively updating patrol policies given fixed dispatch policy and vice versa. Within this framework, we extend previous works addressing combinatorial action selection in RL via an MIP approach, where our novel procedure facilitates action selection in settings with stochastic state transitions. We comprehensively demonstrate that policies learned using this joint optimization approach outperform those optimized for patrol or dispatch alone, in addition to heuristics such as random patrol and priority-queue-based dispatch.

The experiments of Section~\ref{sec:efficient_response_experiment} reveal that the joint optimization procedure is robust to settings with a range of incident arrival dynamics. In the high call volume setting, the jointly-optimized policy performs similarly to the dispatch-only optimized policy. In the low call volume setting, the jointly-optimized policy performs similarly to the patrol-only optimized policy. The jointly-optimized policy is the only policy to perform effectively in both settings. In Section~\ref{sec:equity_experiment}, a modified reward is used to obtain more fair policies than those trained ``greedily''. The results reveal that our procedure is flexible to the reward definition and can generate policies that yield more balanced responses and coverage across groups. While there is still much room for improvement according to the equity-based metrics, this represents a promising direction for further research within the framework of joint policy optimization. Finally, Section~\ref{sec:apd_cfs} suggests that the outlined method applies to real-world scenarios by applying joint policy optimization to a setting based on Southwest Atlanta. A comparison to a realistic policy based on real patroller trajectory data reveals significant improvements in response time distribution characteristics due to policy optimization; particularly, jointly-optimized policies for patrol and dispatch are most effective.

Future work can incorporate additional incident-level information in the decision-making process. For instance, the experiment of Section~\ref{sec:equity_experiment} indicates that the encoding of group attributes in the reward can lead to a more equitable response. As with prior works considering fairness in police patrol and dispatch, incidents can also include information regarding socioeconomic and racial status, going beyond the geographic location-based fairness demonstrated in this work. Algorithmically, future research can also focus on the incorporation of fairness into the learning procedure itself, drawing inspiration from prior work on fairness in learning for contextual bandits~\citep{joseph2016fairness} and RL~\citep{jabbari2017fairness}. Such studies require that certain equity constraints are satisfied throughout the learning process, which could be relevant to online learning of policies for patrol and dispatch.

\section*{Acknowledgement}

This work is partially supported by an NSF CAREER CCF-1650913, NSF DMS-2134037, CMMI-2015787, CMMI-2112533, DMS-1938106, DMS-1830210, and the Coca-Cola Foundation.

\bibliographystyle{msom/informs2014}
\bibliography{main}

\newpage
%


\begin{appendix}

\section{Model Selection}\label{app:model_selection}

\subsection{Model Selection in Section~\ref{sec:efficient_response_experiment}}\label{app:model_selection_section_1}
\begin{figure}[t]
    \centering
    \includegraphics[width=\textwidth]{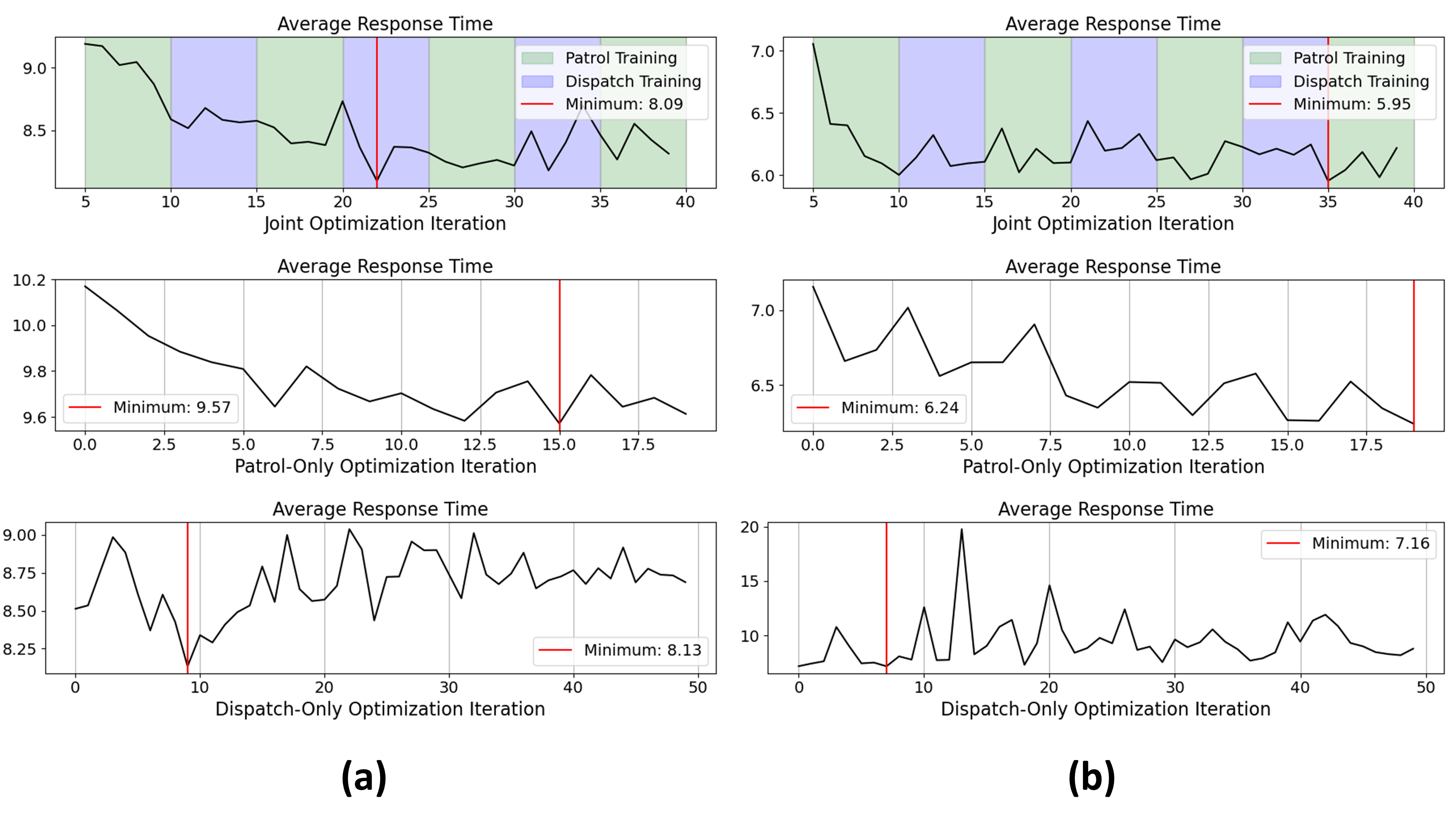}
    \caption{In the experiments of Section~\ref{sec:efficient_response_experiment}, the goal is to generate a policy that efficiently responds to incidents. The average response time is tracked over the course of training using $n_{\rm episodes}=100$ validation simulations per iteration, and the policy with the lowest average response time is selected.}
    \label{fig:model_selection_section_1}
\end{figure}

In the experiments of Section~\ref{sec:efficient_response_experiment}, the desired policy is that which responds to incidents most efficiently. As such, the average response time is tracked over $n_{\rm episodes}=100$ of on-policy validation simulation over the course of optimization for each of the three optimized policies in each setting. This is depicted for the high call volume setting in Figure~\ref{fig:model_selection_section_1}(a), which demonstrates that the policies corresponding to the minimum average response time are the jointly-optimized policy after iteration 22, the patrol-only optimized policy after iteration 15, and the dispatch-only optimized policy after iteration 9. Similarly, in the low call volume setting depicted in Figure~\ref{fig:model_selection_section_1}(b), the selected policies are the jointly-optimized policy after iteration 35, the patrol-only optimized policy after iteration 19, and the dispatch-only optimized policy after iteration 7.

\subsection{Model Selection in Section~\ref{sec:equity_experiment}}
\begin{figure}[t]
    \centering
    \includegraphics[width=\textwidth]{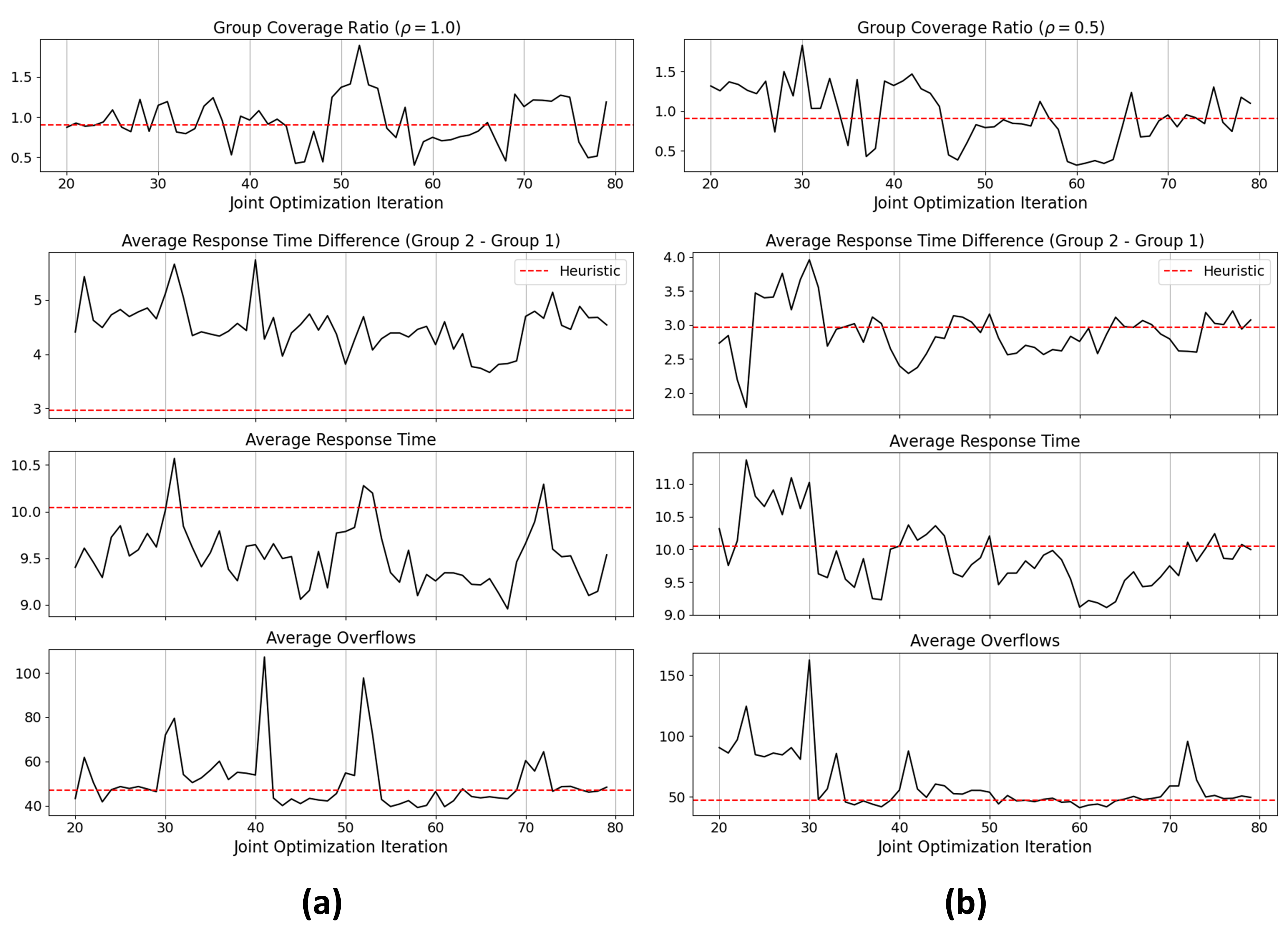}
    \caption{In the experiments of Section~\ref{sec:equity_experiment}, the goal is to balance response time and coverage across groups. These metrics are tracked through training for the ``unfair'' policy in (a) and the ``fair'' policy in (b). In each case, the optimized networks are selected to balance these two metrics (first and second rows), subject to constraints on the effectiveness of the policies (third and fourth rows).}
    \label{fig:model_selection_section_2}
\end{figure}

In the experiment of Section~\ref{sec:equity_experiment}, the goal is to find a policy that minimizes the difference in average response time between the two groups while balancing coverage. The characteristics of the ``unfair'' ($\rho_{\rm large}=1.0$) policy over the course of training are depicted in Figure~\ref{fig:model_selection_section_2}(a), where coverage is computed using $n_{\rm episodes}=25$ on-policy validation simulations in each iteration, and response time characteristics are computed over $n_{\rm episodes}=100$ simulations (the same is true of the ``fair" policy in Figure~\ref{fig:model_selection_section_2}(b)). With the goal of this section in mind, the unfairly selected policy corresponds to iteration 66, which has a minimal average difference in response time and comparable coverage to the heuristic policy. The fair policy is chosen similarly - the goal is to balance fair response and fair coverage while still yielding an ``effective policy''. With this in mind, while iteration 23 corresponds to the most balanced average response in the top panel of Figure~\ref{fig:model_selection_section_2}(b), the policy is not ``effective'' since there is a very large number of incident queue overflows (bottom panel). Therefore, the fair ($\rho_{\rm large}=0.5$) policy used in Section~\ref{sec:equity_experiment} corresponds to the optimized policy after iteration 36, which improves upon the balancing of average response time (second panel) while maintaining few overflows of the incident queue (bottom panel) and yielding a better coverage ratio (top panel).

\subsection{Model Selection in Section~\ref{sec:apd_cfs}}
\begin{figure}[t]
    \centering
    \includegraphics[width=\textwidth]{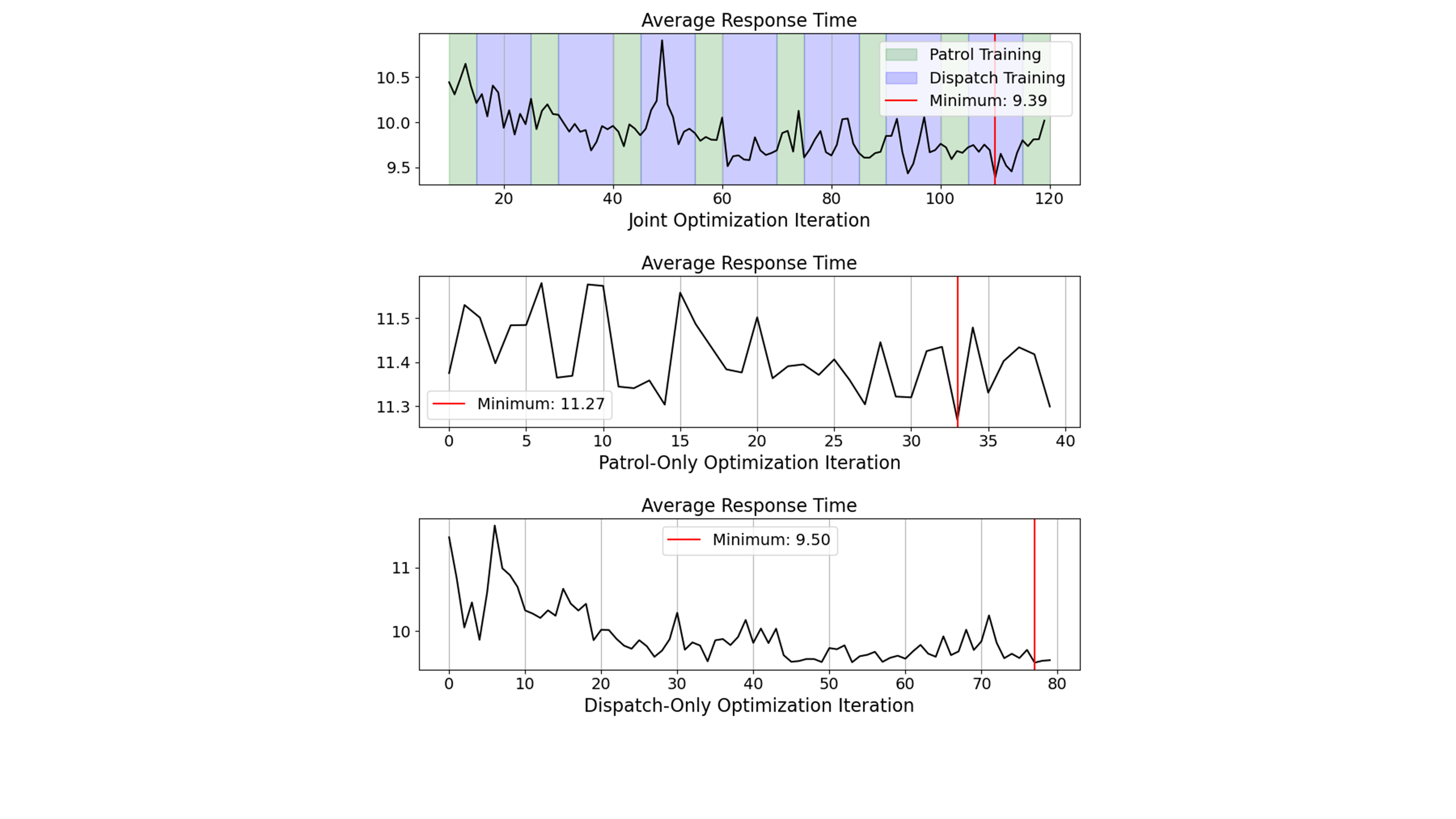}
    \vspace{-0.5in}
    \caption{As with the first set of experiments, in the experiment of Section~\ref{sec:apd_cfs}, the average response time is computed using $n_{\rm episodes}=100$ validation simulations per iteration of training. The fastest-average-response policy is selected, subject to a constraint that the average number of overflows is less than 1.25 times that of the heuristic policy.}
    \label{fig:model_selection_section_3}
\end{figure}

Just as in Section~\ref{sec:efficient_response_experiment} and Appendix~\ref{app:model_selection_section_1}, in the experiment of Section~\ref{sec:apd_cfs}, the policy is selected to minimize the average response time. Therefore, this metric is tracked over $n_{\rm episodes}=100$ on-policy simulated episodes throughout the optimization of each policy. This is visualized in Figure~\ref{fig:model_selection_section_3}, indicating that the optimal policies are those after iteration 110 for the jointly-optimized policy, after iteration 33 for the patrol-only optimized policy, and after iteration 77 for the dispatch-only optimized policy.

\end{appendix}

\end{document}